\documentclass[conference]{IEEEtran}
\IEEEoverridecommandlockouts
\usepackage{cite}
\usepackage{amsmath,amssymb,amsfonts}
\usepackage{algorithm}
\usepackage{algorithmic}
\usepackage{graphicx}
\usepackage{textcomp}
\usepackage{xcolor}
\usepackage{listings}
\usepackage{subcaption}
\usepackage{tcolorbox}
\usepackage{balance}
\usepackage{multirow}
\usepackage{enumitem}

\tcbuselibrary{breakable}

\def\BibTeX{{\rm B\kern-.05em{\sc i\kern-.025em b}\kern-.08em
    T\kern-.1667em\lower.7ex\hbox{E}\kern-.125emX}}

\usepackage[acronym,toc,shortcuts]{glossaries}

\newacronym{3GPP}{3GPP}{3rd Generation Partnership Project }
\newacronym{5G}{5G}{Fifth Generation}
\newacronym{AAA}{AAA}{Authentication, Authorization and Accounting}
\newacronym{AGV}{AGV}{Automated Guided Vehicle}
\newacronym{AI}{AI}{Artificial Intelligence}
\newacronym{AP}{AP}{Access Point}
\newacronym{API}{API}{Application Programming Interface}
\newacronym{AMF}{AMF}{Access and Mobility Management Function}
\newacronym{APN}{APN}{Access Point Name}
\newacronym{AR}{AR}{Augmented Reality}
\newacronym{ANR}{ANR}{Automatic Neighbour Relationship}
\newacronym{ASN}{ASN}{Abstract Syntax Notation}
\newacronym{AWS}{AWS}{Amazon Web Services}
\newacronym{BaaS}{BaaS}{Backend-as-a-Service}
\newacronym{BS}{BS}{Base Station}
\newacronym{BB}{BB}{BaseBand}
\newacronym{BBU}{BBU}{BaseBand Unit}
\newacronym{BCN}{BCN}{Blockchain Network}
\newacronym{BER}{BER}{Bit Error Rate}
\newacronym{BSSID}{BSSID}{Basic Service Set Identification}
\newacronym{CaaS}{CaaS}{Containers as a Service}
\newacronym{CaPEX}{CaPEX}{Capital Expenditure}
\newacronym{ClaaS}{ClaaS}{Cloud as a Service}
\newacronym{CA}{CA}{Carrier Aggregation}
\newacronym{CAT}{CAT}{Capacity-Aware TOPSIS}
\newacronym{CAPEX}{CAPEX}{Capital Expenditure}
\newacronym{CD}{CD}{Continuous Development}
\newacronym{CDN}{CDN}{Content Delivery Network}
\newacronym{CEP}{CEP}{Complex Event Processing}
\newacronym{CELL-ID}{CELL-ID}{cell identification ID}
\newacronym{CGI}{CGI}{Cell Global Identification}
\newacronym{CI}{CI}{Continuous Integration}
\newacronym{CLSM}{CLSM}{Closed loop spatial multiplexing}
\newacronym{CQI}{CQI}{Channel Quality Indicator}
\newacronym{CN}{CN}{Core Network}
\newacronym{CNF}{CNF}{Cloud-native Network Function}
\newacronym{CoMP}{CoMP}{Coordinated Multipoint Transmission/Reception}
\newacronym{CP}{CP}{Connection Point}
\newacronym{CPS}{CPS}{Cyber-Physical System}
\newacronym{CPU}{CPU}{Central Processing Unit}
\newacronym{CNA}{CNA}{Cloud-Native Application}
\newacronym{CNFC}{CNFC}{CNF Component}
\newacronym{CNCF}{CNCF}{Cloud Native Computing Foundation}
\newacronym{CNFVI}{CNFVI}{Cloud-native NFV}
\newacronym{CNN}{CNN}{Convolutional Neural Networks}
\newacronym{CNTT}{CNTT}{Cloud INfrastructure Telco Taskforce}
\newacronym{CL}{CL}{Closed-Loop}
\newacronym{CSMF}{CSMF}{Core Slice Management Function}
\newacronym{CU}{CU}{Central Unit}
\newacronym{COTS}{COTS}{Commercial off-the-shelf}
\newacronym{CPRI}{CPRI}{Common Public Radio Interface}
\newacronym{CRM}{CRM}{Customer Relationship Management}
\newacronym{CS}{CS}{central scheduler}
\newacronym{CSI}{CSI}{Channel Status Information}
\newacronym{CSP}{CSP}{Communication Service Provider}
\newacronym{eNB}{eNB}{evolved Node-B}
\newacronym{D2D}{D2D}{Device-to-Device}
\newacronym{DC}{DC}{Data Center}
\newacronym{DCN}{DCN}{Dedicated Core Network}
\newacronym{DCS}{DCS}{Distributed Control System}
\newacronym{DECOR}{DECOR}{Dedicated Core Network}
\newacronym{DevOps}{DevOps}{Development and Operations}
\newacronym{DTLS}{DTLS}{Datagram Transport Layer Security}
\newacronym{DL}{DL}{Downlink}
\newacronym{DLT}{DLT}{Distributed Ledger Technology}
\newacronym{DID}{DID}{Decentralized Identifier}
\newacronym{DNS}{DNS}{Domain Name Server}
\newacronym{DRL}{DRL}{Deep Reinforcement Learning}
\newacronym{DU}{DU}{Distributed Unit}
\newacronym{DMM}{DMM}{Distributed Mobility Management}
\newacronym{DPU}{DPU}{Data Processing Unit}
\newacronym{DPDK}{DPDK}{Data Plane Development Kit}
\newacronym{eBPF}{eBPF}{extended Berkeley Packet Filter}
\newacronym{eCPRI}{eCPRI}{enhanced CPRI}
\newacronym{ECA}{ECA}{Event-Condition-Action}
\newacronym{eMBB}{eMBB}{enhanced Mobile Broadband}
\newacronym{EMI}{EMI}{Electromagnetic Interference}
\newacronym{EMC}{EMC}{Electromagnetic Compatibility}
\newacronym{eNodeB}{eNodeB}{evolved Node-B}
\newacronym{E2E}{E2E}{End-to-End}
\newacronym{EPC}{EPC}{Evolved Packet Core}
\newacronym{EPS}{EPS}{Evolved Packet System}
\newacronym{E-RAB}{E-RAB}{E-UTRAN Radio Access Bearer}
\newacronym{ERP}{ERP}{Enterprise Resource Planning}
\newacronym{ETSI}{ETSI}{European Telecommunications Standards Institute}
\newacronym{FaaS}{FaaS}{Functions as a Service}
\newacronym{FCC}{FCC}{Federal Communications Commission}
\newacronym{FDD}{FDD}{Frequency Division Duplexing}
\newacronym{FEM}{FEM}{Flow Extraction Manager}
\newacronym{FoF}{FoF}{Factories-of-the-Future}
\newacronym{FMC}{FMC}{Fixed Mobile Convergence}
\newacronym{FPGA}{FPGA}{Field-Programmable Gate Array}
\newacronym{FTP}{FTP}{File Transfer Protocol}
\newacronym{FCAPS}{FCAPS}{Fault, Configuration, Accounting, Performance and Security}
\newacronym{GCP}{GCP}{Google Cloud Platform}
\newacronym{GDPR}{GDPR}{General Data Protection Regulation}
\newacronym{GGSN}{GGSN}{Gateway GPRS Support Node}
\newacronym{GPRS}{GPRS}{General packet radio service}
\newacronym{GPS}{GPS}{Global Positioning System}
\newacronym{gRPC}{gRPC}{gRPC Remote Procedure Calls}
\newacronym{GTP}{GTP}{GPRS Tunneling Protocol}
\newacronym{HetNet}{HetNet}{heterogeneous network}
\newacronym{HO}{HO}{Handover}
\newacronym{HRoT}{HRoT}{Hardware Root of Trust}
\newacronym{HSS}{HSS}{Home Subscriber Station}
\newacronym{HTTP}{HTTP}{Hypertext Transfer Protocol}
\newacronym{HD}{HD}{High-Definition}
\newacronym{HDFS}{HDFS}{Hadoop Distributed File System}
\newacronym{HiveQL}{HiveQL}{Hive Query language}
\newacronym{HMI}{HMI}{Human-Machine Interface}
\newacronym{HNSM}{HNSM}{Host-based Network Security Manager}
\newacronym{HSPA}{HSPA}{High Speed Packet Access}
\newacronym{HSM}{HSM}{Hardware Security Module}
\newacronym{IAB}{IAB}{Integrated Access and Backhaul}
\newacronym{IaC}{IaC}{Infrastructure as Code}
\newacronym{IaaS}{IaaS}{Infrastructure as a Service}
\newacronym{IBLER}{IBLER}{Initial Block Error Rate}
\newacronym{ICIC}{ICIC}{inter-cell interference coordination}
\newacronym{ICN}{ICN}{information-centric network}
\newacronym{ICT}{ICT}{Information and Communication Technologies}
\newacronym{IEC}{IEC}{International Electrotechnical Commission}
\newacronym{IEEE}{IEEE}{Institute of Electrical and Electronics Engineers}
\newacronym{IETF}{IETF}{Internet Engineering Task Force}
\newacronym{IGA}{IGA}{Identity Governance and Administration}
\newacronym{IIoT}{IIoT}{Industrial IoT}
\newacronym{IMSI}{IMSI}{International Mobile Subscriber Identity}
\newacronym{IMEI}{IMEI}{International Mobile Station Equipment Identity}
\newacronym{IMS}{IMS}{IP Multimedia Subsystem}
\newacronym{IDoT}{IDoT}{Identity of Things}
\newacronym{ID}{ID}{Identifier}
\newacronym{I/O}{I/O}{Input/Output}
\newacronym{ICMP}{ICMP}{Internet Control Message Protocol}
\newacronym{ISA}{ISA}{International Society of Automation}
\newacronym{IoT}{IoT}{Internet of Things}
\newacronym{IP}{IP}{Internet Protocol}
\newacronym{IRS}{IRS}{Intelligent Reflective Surface}
\newacronym{ISO}{ISO}{International Organization for Standardization}
\newacronym{ITU}{ITU}{International Telecommunication Union}
\newacronym{IT}{IT}{Information Technology}
\newacronym{ITS}{ITS}{Intelligent Transportation Systems}
\newacronym{Li-Fi}{Li-Fi}{Light Fidelity}
\newacronym{L2}{L2}{Layer 2}
\newacronym{LBO}{LBO}{Local Break Out}
\newacronym{LCM}{LCM}{Life Cycle Management}
\newacronym{LDPC}{LDPC}{Low Density Parity Check}
\newacronym{LPWAN}{LPWAN}{Low Power Wide Area Network}
\newacronym{L3}{L3}{Layer 3}
\newacronym{GBR}{GBR}{Guaranteed Bit Rate}
\newacronym{GLUE}{GLUE}{General Language Understanding Evaluation}
\newacronym{MI}{MI}{Middleware Interface}
\newacronym{JSON}{JSON}{JavaScript Object Notation}
\newacronym{IMT}{IMT}{International Mobile Telecommunications}
\newacronym{K8s}{K8s}{Kubernetes}
\newacronym{KMS}{KMS}{Key Management Service}
\newacronym{KPI}{KPI}{Key Performance Indicator}
\newacronym{LA}{LA}{Location Area}
\newacronym{LAC}{LAC}{location area code}
\newacronym{LDAP}{LDAP}{Lightweight Directory Access Protocol}
\newacronym{LMA}{LMA}{Local Mobility Anchor}
\newacronym{LTE}{LTE}{long term evolution}
\newacronym{mMTC}{mMTC}{massive Machine-Type Communications}
\newacronym{MADM}{MADM}{Multiple Attribute Decision Making}
\newacronym{MCC}{MCC}{Mobile Country Code}
\newacronym{MCS}{MCS}{Modulation Coding Scheme}
\newacronym{MNC}{MNC}{Mobile Network Code}
\newacronym{MIMO}{MIMO}{multiple-input multiple-output}
\newacronym{MAG}{MAG}{Mobile Access Gateway}
\newacronym{MAAR}{MAAR}{Mobility Anchor and Access Router}
\newacronym{MANO}{MANO}{Management and Orchestration}
\newacronym{MEC}{MEC}{Multi-access edge computing}
\newacronym{MFA}{MFA}{Multi-Factor Authentication}
\newacronym{ML}{ML}{Machine Learning}
\newacronym{MME}{MME}{Mobility Management Entity}
\newacronym{MDM}{MDM}{Mobile Device Management}
\newacronym{MN}{MN}{Mobile Node}
\newacronym{MTC}{MTC}{machine Type Communication}
\newacronym{MNO}{MNO}{Mobile Network Operator}
\newacronym{MR}{MR}{Mixed Reality}
\newacronym{MSISDN}{MSISDN}{Mobile Station International Subscriber Directory Number}
\newacronym{NAT}{NAT}{Network Address Translation}
\newacronym{NBI}{NBI}{NorthBound Interface}
\newacronym{NC}{NC}{Network Coding}
\newacronym{NEF}{NEF}{Network Exposure Function}
\newacronym{NF}{NF}{Network Function}
\newacronym{NFV}{NFV}{Network Functions Virtualization}
\newacronym{NFVO}{NFVO}{Network Functions Virtualization Orchestrator}
\newacronym{NFVI}{NFVI}{Network Functions Virtualization Infrastructure}
\newacronym{NIST}{NIST}{National Institute of Standards and Technology}
\newacronym{NIC}{NIC}{Network Interface Card}
\newacronym{NLP}{NLP}{Natural Language Processing}
\newacronym{NLU}{NLU}{Natural Language Understanding}
\newacronym{NOMA}{NOMA}{Non-Orthogonal Multiple Access}
\newacronym{NoSQL}{NoSQL}{Not Only SQL}
\newacronym{NPN}{NPN}{Non-Public Network}
\newacronym{NR}{NR}{New Radio}
\newacronym{NRF}{NRF}{Network Repository Function}
\newacronym{NS}{NS}{Network Service}
\newacronym{NSSMF}{NSSMF}{Network Slice Subnet Management Function}
\newacronym{NSMF}{NSMF}{Network Slice Selection Function}
\newacronym{NTN}{NTN}{Non-Terrestrial Networks}
\newacronym{QoS}{QoS}{Quality-of-Service}
\newacronym{QoE}{QoE}{Quality-of-Experience}
\newacronym{PaaS}{PaaS}{Platform as a Service}
\newacronym{PoP}{PoP}{Point of Presence}
\newacronym{PNF}{PNF}{Physical Network Function}
\newacronym{PCB}{PCB}{Printed Circuit Board}
\newacronym{PDN}{PDN}{packet data network}
\newacronym{PI}{PI}{Provision Interface}
\newacronym{PID}{PID}{Proportional Integral Derivative}
\newacronym{PF}{PF}{Proportional Fair}
\newacronym{P-GW}{P-GW}{packet gateway}
\newacronym{PDP}{PDP}{Packet Data Protocol}
\newacronym{PDU}{PDU}{Packet Data Unit}
\newacronym{PFH}{PFH}{Probability of Failure on Demand per Hour}
\newacronym{PHY}{PHY}{physical layer}
\newacronym{PKI}{PKI}{Public Key Infrastructure}
\newacronym{PLC}{PLC}{Programmable Logic Controller}
\newacronym{PLMN}{PLMN}{Public Land Mobile Network}
\newacronym{PMIPv6}{PMIPv6}{Proxy Mobile IPv6}
\newacronym{PMI}{PMI}{Precoding Matrix Index}
\newacronym{PQC}{PQC}{Post Quantum Cryptography}
\newacronym{PRB}{PRB}{Physical Resource Block}
\newacronym{PUSCH}{PUSCH}{Physical Uplink Shared Channel}
\newacronym{QAM}{QAM}{Quadrature amplitude modulation}
\newacronym{QCI}{QCI}{QoS Class Identifier}
\newacronym{QKD}{QKD}{Quantum Key Distribution}
\newacronym{RA}{RA}{Routing Area}
\newacronym{RAMI}{RAMI}{Reference Architectural Model Industrie}
\newacronym{RB}{RB}{Resource Block}
\newacronym{REST}{REST}{Representational State Transfer}
\newacronym{RPC}{RPC}{Remote Procedure Call}
\newacronym{RI}{RI}{Rank Indicator}
\newacronym{RAN}{RAN}{Radio Access Network}
\newacronym{RBAC}{RBAC}{Role-Based Access Control}
\newacronym{RFC}{RFC}{Request for Comment}
\newacronym{RF}{RF}{Radio Frequency}
\newacronym{RFID}{RFID}{Radio-frequency identification}
\newacronym{RRC}{RRC}{Radio Resource Control}
\newacronym{RRU}{RRU}{Remote Radio Unit}
\newacronym{RNC}{RNC}{radio network controller}
\newacronym{RNN}{RNN}{Recurrent Neural Networks}
\newacronym{RSSI}{RSSI}{Received Signal Strength Indicator}
\newacronym{RSRP}{RSRP}{Reference Signal Received Power}
\newacronym{RTT}{RTT}{Round Trip Time}
\newacronym{OSI}{OSI}{Open Systems Interconnection}
\newacronym{OPEX}{OPEX}{Operational Expenditure}
\newacronym{OAM}{OAM}{Operation, Administration and Management}
\newacronym{OL}{OL}{Open-Loop}
\newacronym{ONAP}{ONAP}{Open Networking Automation Platform}
\newacronym{ONF}{ONF}{Open Networking Foundation}
\newacronym{ONOS}{ONOS}{Open Network Operating System}
\newacronym{O-RAN}{O-RAN}{Open Radio Access Network}
\newacronym{OpEX}{OpPEX}{Operational Expenditure}
\newacronym{OPNFV}{OPNFV}{Open Platform for Network Functions Virtualization}
\newacronym{OS}{OS}{Operating System}
\newacronym{OSM}{OSM}{Open Source MANO}
\newacronym{OSS}{OSS}{Operational Support Systems}
\newacronym{OT}{OT}{Operational Technology}
\newacronym{OVP}{OVP}{OPNFV Verification Program}
\newacronym{OTT}{OTT}{over-the-top}
\newacronym{RoT}{RoT}{Root of Trust}
\newacronym{SaaS}{SaaS}{Software as a Service}
\newacronym{SA}{SA}{Stand Alone}
\newacronym{SAML}{SAML}{Security Assertion Markup Language}
\newacronym{SB}{SB}{Service Bus}
\newacronym{SI}{SI}{Service Interface}
\newacronym{SIL}{SIL}{Safety Integrity Level}
\newacronym{SAC}{SAC}{service area code}
\newacronym{SBA}{SBA}{Service Based Architecture}
\newacronym{SBI}{SBI}{Service Based Interface}
\newacronym{SD}{SD}{service Discovery}
\newacronym{SCMA}{SCMA}{Sparse Code Multiple Access}
\newacronym{SDLC}{SDLC}{Software Development Life Cycle}
\newacronym{SFC}{SFC}{Service Function Chaining}
\newacronym{SMI}{SMI}{Service Management Interface}
\newacronym{SLA}{SLA}{Service Level Agreement}
\newacronym{SDN}{SDN}{Software Defined Networking}
\newacronym{SDK}{SDK}{Software Development Kit}
\newacronym{SDO}{SDO}{Standards Developing Organization}
\newacronym{SOA}{SOA}{Service Oriented Architecture}
\newacronym{SFN}{SFN}{Single Frequency Network}
\newacronym{SEI}{SEI}{Service External Interface}
\newacronym{SMW}{SMW}{Software Middleware}
\newacronym{S-GW}{S-GW}{serving gateway}
\newacronym{SR-IOV}{SR-IOV}{Single Root Input Output Virtualization}
\newacronym{SINR}{SINR}{signal-to-interference-plus-noise ratio}
\newacronym{SGSN}{SGSN}{Serving GPRS Support Node}
\newacronym{SSI}{SSI}{Self Sovereign Identity}
\newacronym{SSID}{SSID}{Service Set Identification}
\newacronym{SSL}{SSL}{Secure Socket Layer}
\newacronym{STO}{STO}{Safe Torque Off}
\newacronym{SVD}{SVD}{singular value decomposition}
\newacronym{SW}{SW}{Software}
\newacronym{SU}{SU}{Service Unit}
\newacronym{TAR}{TAR}{Tape Archive}
\newacronym{TCP}{TCP}{transport control protocol}
\newacronym{TLS}{TLS}{Transport Layer Security}
\newacronym{TCO}{TCO}{total cost ownership}
\newacronym{TDD}{TDD}{Time Division Duplexing}
\newacronym{TM}{TM}{transmission mode}
\newacronym{TSN}{TSN}{Time Sensitive Network}
\newacronym{TUG}{TUG}{Telecom User Group}
\newacronym{TEID}{TEID}{tunnel endpoint identifier}
\newacronym{UAV}{UAV}{Unmanned Aerial Vehicle}
\newacronym{UDN}{UDN}{Ultra Dense Network}
\newacronym{UMTS}{UMTS}{Universal Mobile Telecommunications Service} 
\newacronym{UE}{UE}{user equipment}
\newacronym{UL}{UL}{Uplink}
\newacronym{UP}{UP}{User Plane}
\newacronym{URL}{URL}{Uniform Resource Locator}
\newacronym{uRLLC}{uRLLC}{ultra-Reliable Low-Latency Communication}
\newacronym{UDP}{UDP}{User Datagram Protocol}
\newacronym{UPF}{UPF}{User Plane Function}
\newacronym{VC}{VC}{Verifiable Credential}
\newacronym{VIM}{VIM}{Virtual Infrastructure Manager}
\newacronym{VL}{VL}{Virtual Link}
\newacronym{VF}{VF}{Virtual Function}
\newacronym{VNF}{VNF}{Virtual Network Function}
\newacronym{VPN}{VPN}{Virtual Private Network}
\newacronym{vRAN}{vRAN}{Virtual Radio Access Network}
\newacronym{VNFC}{VNFC}{Virtual Network Function Component}
\newacronym{VNFM}{VNFM}{Virtual Network Function Manager}
\newacronym{VNFFG}{VNFFG}{Virtualized Network Function Forwarding Graphs}
\newacronym{VM}{VM}{Virtual Machine}
\newacronym{vRC}{vRC}{Virtual Radio Controller}
\newacronym{vPP}{vPP}{Virtual Packet Processor}
\newacronym{VR}{VR}{Virtual Reality}
\newacronym{XR}{XR}{Extended Reality}
\newacronym{WiFi}{WiFi}{Wireless Fidelity}
\newacronym{WLAN}{WLAN}{Wireless Local Area Network}
\newacronym{ZSM}{ZSM}{Zero touch network \& Service Management}
\newacronym{ZKP}{ZKP}{Zero-Knowledge Proof}

\begin{document}

\title{F-KANs: Federated Kolmogorov-Arnold Networks
}

\author{Engin~Zeydan$^{\ast}$, Cristian J. Vaca-Rubio$^{\ast}$, Luis~Blanco$^{\ast}$, Roberto Pereira$^{\ast}$, Marius Caus$^{\ast}$, Abdullah Aydeger$^{\dagger}$\\% <-this % stops a space
$^{\ast}$Centre Tecnològic de Telecomunicacions de Catalunya (CTTC), Barcelona, Spain, 08860.\\
$^{\dagger}$ Dept. of  Electrical Engineering and Computer Science, Florida Institute of Technology, Melbourne, FL, USA \\
\protect Email: \{ezeydan, cvaca, lblanco, rpereira, mcaus\}@cttc.cat, aaydeger@fit.edu.
\thanks{This work was partially funded by “ERDF A way of making Europe” project PID2021-126431OB-I00 and Spanish MINECO - Program UNICO I+D (grants TSI-063000-2021-54 and -55) Grant PID2021-126431OB-I00 funded by MCIN/AEI/ 10.13039/501100011033 and 5G-STARDUST project  under Grant Agreement No 101096573.}}
\maketitle

\begin{abstract}
In this paper, we present an innovative federated learning (FL) approach that utilizes Kolmogorov-Arnold Networks (KANs) for classification tasks. By utilizing the adaptive activation capabilities of KANs in a federated framework, we aim to improve classification capabilities while preserving privacy. The study evaluates the performance of federated KANs (F-KANs) compared to traditional federated Multi-Layer Perceptrons (F-MLPs)  on  classification task.  The results show that the F-KANs model significantly outperforms the F-MLP model in terms of accuracy, precision, recall, F1 score and stability, and achieves better performance, paving the way for more efficient and privacy-preserving predictive analytics.
\end{abstract}


\section{Introduction}

Classification tasks are of central importance for various applications such as medical diagnosis, spam detection and image recognition \cite{b1}. Traditional methods often rely on large centralized datasets, which raises concerns about data privacy and potential data breaches. Machine learning models, particularly deep learning architectures, have shown promise when it comes to capturing complicated patterns for classification \cite{b2}. However, these models often require large amounts of data, which poses privacy and data security issues. Federated Learning (FL) provides a solution by enabling collaborative model training across multiple decentralized devices while keeping the data localized \cite{b3}.  On the other hand, research on the application of KANs in wireless networks is a relatively new and specialized area within the broader context of neural networks and their applications to wireless communications.  The paper \cite{b11} presents a novel method to improve intrusion detection in Internet of Things (IoT) environments where the proposed approach leverages KANs in combination with ensemble learning techniques to enhance the accuracy and robustness of intrusion detection systems (IDS). The paper \cite{b12} introduces an explainable unsupervised anomaly detection framework specifically designed for the Industrial Internet of Things (IIoT) and uses Convolutional and CBAM-enhanced Temporal Attention KAN (CCTAK). However, the application of federated learning in this context has not yet been presented.

In this study, federated KANs (F-KANs) are presented that combine the strengths of FL with the innovative architecture of KANs \cite{b4} inspired by the Kolmogorov-Arnold representation theorem. In contrast to conventional neural networks, KANs use spline-based univariate functions as learnable activation functions, which improves their adaptability and interpretability. Previous works on KANs have  explored the extension of KANs into convolutional architectures \cite{b5}, graph-structured data \cite{b6}, time series prediction \cite{b7, b8}, vision tasks \cite{b9}, GNN model leveraging KANs \cite{b10}. Despite these advances, the application of KANs in federated learning (referred to as F-KANs) has not yet been explored. This paper investigates the application of F-KANs to classification tasks and compares their performance with traditional F-MLPs on a diverse classification dataset. Table \ref{tab:notation} contains the notation table used throughout the paper.

\begin{table}[h!]
\scriptsize
\centering
\caption{Notations used throughout the paper.}
\begin{tabular}{|c|p{6cm}|}
\hline
\textbf{Notation} & \textbf{Description} \\
\hline
$(X, y)$ & Dataset features and labels \\
$(X_{train}, y_{train})$ & Training set features and labels \\
$(X_{test}, y_{test})$ & Testing set features and labels \\
$\mathcal{D}_{train}$ & Training TensorDataset \\
$\mathcal{D}_{test}$ & Testing TensorDataset \\
$N$ & Number of clients \\
$\{\mathcal{D}_{train}^i\}_{i=1}^N$ & Partitioned training datasets for each client \\
$\theta_G$ & Global model parameters \\
$\{\theta_i\}_{i=1}^N$ & Client model parameters \\
$\mathcal{L}$ & Loss \\
$\mathcal{A}$ & Accuracy \\
$R$ & Number of federated learning rounds \\
$S$ & Number of training steps per round \\
%$N_{\text{in}}$ & Number of inputs to a KAN layer \\
%$N_{\text{out}}$ & Number of outputs from a KAN layer \\
$\Phi$ & Matrix comprising univariate functions in a KAN layer \\
$\varphi_{i,j}(\cdot)$ & Univariate function from input $i$ to output $j$ in a KAN layer \\
$n_k$ & Number of data samples in client $k$ \\
$m_r$ & Total number of data samples for round $r$ \\
%$\mathcal{C}$ & Set of clients\\
\hline
\end{tabular}
\label{tab:notation}
%\vspace{-.5cm}
\end{table}

\section{Problem Statement and Federated Kolmogorov-Arnold Networks Background}

We address  the problem of classification in a federated learning environment. Specifically, we are concerned with classifying data points based on features that are distributed across multiple clients. Each data point at time $t$ is represented by a feature vector $\textbf{x}_t$ and the goal is to predict the class label $y_t$. The Kolmogorov-Arnold representation theorem underlies the architecture of KANs and allows any multivariate continuous function to be represented as a composition of simple univariate functions. This study extends this concept to a federated environment where each client trains a F-KAN model locally and a central server aggregates the model updates without accessing the raw data, thus preserving privacy.

Federated Kolmogorov-Arnold Networks (F-KANs) extend the concept of KANs by leveraging the Kolmogorov-Arnold representation theorem within a FL framework. The theorem states that any multivariate continuous function can be represented by a finite composition of univariate functions. This allows F-KANs to replace traditional linear weights with spline-based univariate functions along the edges of the network, structured as learnable activation functions, while enabling decentralized training across multiple clients. An F-KAN layer is defined by a matrix $\Phi$ comprising univariate functions ${\varphi_{i,j}(\cdot)}$ with $i=1,\ldots,N_{\text{in}}$ and $j=1,\ldots,N_{\text{out}}$, where $N_{\text{in}}$ and $N_{\text{out}}$ denote the number of inputs and outputs, respectively. These univariate functions are usually approximated with B-splines, which enable smooth and continuous interpolation between control points. The architecture of F-KANs includes multiple layers, with each layer transforming the input through spline-based activation functions. This design improves the model's ability to learn complex patterns  than traditional federated MLPs (F-MLPs). In a federated environment, multiple clients independently train local F-KAN models on their respective datasets. The local models regularly exchange their updates with a central server, which performs federated averaging to update the global model. This iterative process continues until the global model converges. This ensures better performance while maintaining data privacy between clients.

%with fewer parameters
%KANs utilize the Kolmogorov-Arnold representation theorem, which states that any multivariate continuous function can be represented by a finite composition of univariate functions. This theorem allows KANs to replace traditional linear weights with spline-based univariate functions along the edges of the network, structured as learnable activation functions.
%A KAN layer is defined by a matrix $\Phi$ comprising univariate functions $\{\varphi_{i,j}(\cdot)\}$ with $i=1,\ldots,N_{\text{in}}$ and $j=1,\ldots,N_{\text{out}}$, where $N_{\text{in}}$ and $N_{\text{out}}$ denote the number of inputs and outputs, respectively. These univariate functions are usually approximated with B-splines, which enable smooth and continuous interpolation between control points.
%The generalized architecture of KANs includes multiple layers, with each layer transforming the input through spline-based activation functions. This design improves the model's ability to learn complex patterns with fewer parameters than traditional MLPs.

\section{Proposed Method}

\begin{figure}[htp]
    \centering
    \includegraphics[width=.95\linewidth]{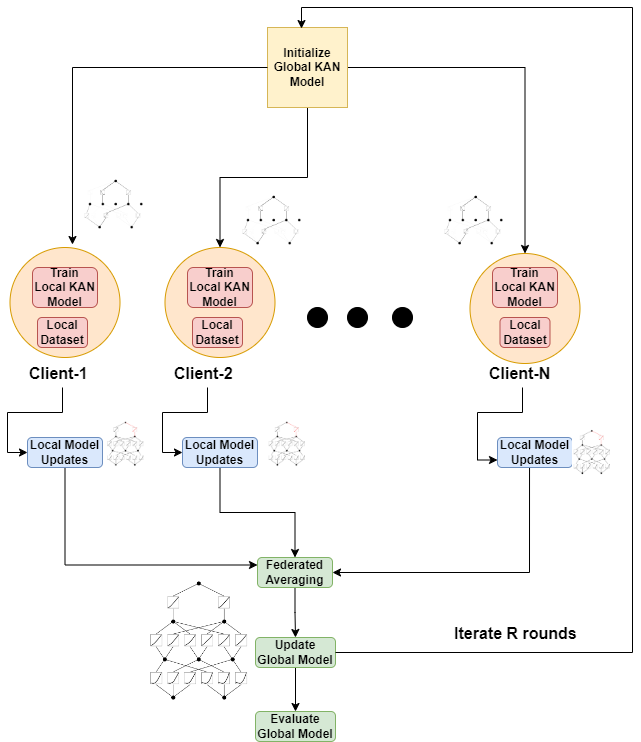}
    \caption{Federated KAN Methodology}
    \label{fig:Federated_KANs}
    \vspace{-.5cm}
\end{figure}

Fig. \ref{fig:Federated_KANs} illustrates the functionality of F-KANs and the interaction between the global model and several clients. At the beginning, the global KAN model is initialized. Each client has its share of the local data. Each client trains a local KAN model using its local data set and sends the model updates back to the central server. The central server performs federated averaging to aggregate the local updates and update the global model. This process is repeated for a certain number of rounds, continuously improving the global model. Finally, the updated global model is evaluated on a test dataset. Fig. \ref{fig:Federated_KANs} shows these steps with nodes representing the main operations and arrows indicating the flow and iteration of the federated learning process. In summary, F-KANs uses decentralized data to jointly train a robust global model without sharing local data directly.  Algorithm \ref{tab:pseudo_code} provides the pseudo-code of the F-KAN algorithm. The main algorithm comprises loading and preprocessing data, partitioning data for clients, defining and initializing the F-KAN classifier and calling the federated learning KAN process.  The individual function definitions in this algorithm are as follows:
$create\_dataset$ converts a data loader into tensors for training input and labels. $fed\_avg$ averages the parameters of client models to update the global model. $compute\_metrics$ computes the average loss, accuracy, precision, recall, and F1 score over a dataset. $federated\_learning\_KAN$ manages the federated learning process by iterating over rounds, training client models, updating the global model, and computing metrics.

\begin{algorithm} [htp!]
\footnotesize
\caption{Federated Learning with KAN}
\begin{algorithmic}[1]

\begin{tcolorbox}[title=Inputs and Outputs, breakable]
\STATE \textbf{Input:} Dataset $(X, y)$, number of clients $N$, number of rounds $R$, steps per round $S$, KAN model architecture
\STATE \textbf{Output:} Trained global model $\theta_G$, training and testing losses $\mathcal{L}_{train}$, $\mathcal{L}_{test}$, training and testing accuracies $\mathcal{A}_{train}$, $\mathcal{A}_{test}$, training and testing precision $\mathcal{P}_{train}$, $\mathcal{P}_{test}$, training and testing recall $\mathcal{R}_{train}$, $\mathcal{R}_{test}$, training and testing F1-scores $\mathcal{F}_{train}$, $\mathcal{F}_{test}$
\end{tcolorbox}

\begin{tcolorbox}[title=Function Definitions]

\STATE \textbf{function} \texttt{create\_dataset}(data\_loader):
\STATE \hspace{1em} Concatenate batches from data\_loader into tensors $\mathbf{X}$, $\mathbf{y}$
\STATE \hspace{1em} \textbf{return} \{train\_input: $\mathbf{X}$, train\_label: $\mathbf{y}$\}

\STATE \textbf{function} \texttt{fed\_avg}($\theta_G, \{\theta_i\}, \{n_i\}$):
\STATE \hspace{1em} Initialize $\theta_G \leftarrow 0$
\STATE \hspace{1em} $m_r \leftarrow \sum_{k \in \{1, \ldots, N\}} n_k$
\STATE \hspace{1em} \textbf{for each} $\theta_i$ \textbf{in} client models:
\STATE \hspace{2em} \textbf{for each} parameter $\theta_G^j$ \textbf{in} $\theta_G$:
\STATE \hspace{3em} $\theta_G^j \leftarrow \theta_G^j + \frac{n_i}{m_r} \theta_i^j$
\STATE \hspace{1em} \textbf{return} $\theta_G$

\STATE \textbf{function} \texttt{compute\_metrics}($\theta, \mathcal{D}$):
\STATE \hspace{1em} Compute $\mathcal{L}$, $\mathcal{A}$, $\mathcal{P}$, $\mathcal{R}$, and $\mathcal{F}$ over $\mathcal{D}$
\STATE \hspace{1em} \textbf{return} $\mathcal{L}$, $\mathcal{A}$, $\mathcal{P}$, $\mathcal{R}$, $\mathcal{F}$

\STATE \textbf{function} \texttt{federated\_learning\_KAN} \\ ($\theta_G, \{\mathcal{D}^i\}, \mathcal{D}_{test}, R, S$):
\STATE \hspace{1em} Store training and testing metrics: $\mathcal{L}_{train}$, $\mathcal{A}_{train}$, $\mathcal{P}_{train}$, $\mathcal{R}_{train}$, $\mathcal{F}_{train}$, $\mathcal{L}_{test}$, $\mathcal{A}_{test}$, $\mathcal{P}_{test}$, $\mathcal{R}_{test}$, $\mathcal{F}_{test}$
\STATE \hspace{1em} \textbf{for} $r = 1$ \textbf{to} $R$:
\STATE \hspace{2em} Create client models $\{\theta_i\}$ from $\theta_G$
\STATE \hspace{2em} \textbf{for} $i = 1$ \textbf{to} $N$:
\STATE \hspace{3em} $\mathcal{D}_i \leftarrow \texttt{create\_dataset}(\mathcal{D}^i)$
\STATE \hspace{3em} Train $\theta_i$ on $\mathcal{D}_i$ for $S$ steps
\STATE \hspace{2em} $\theta_G \leftarrow \texttt{fed\_avg}(\theta_G, \{\theta_i\}, \{n_i\})$
\STATE \hspace{2em} $\mathcal{L}_{train}[r]$, $\mathcal{A}_{train}[r]$, $\mathcal{P}_{train}[r]$, $\mathcal{R}_{train}[r]$, $\mathcal{F}_{train}[r] \leftarrow \texttt{compute\_metrics}(\theta_G, \mathcal{D}_{train})$
\STATE \hspace{2em} $\mathcal{L}_{test}[r]$, $\mathcal{A}_{test}[r]$, $\mathcal{P}_{test}[r]$, $\mathcal{R}_{test}[r]$, $\mathcal{F}_{test}[r] \leftarrow \texttt{compute\_metrics}(\theta_G, \mathcal{D}_{test})$
\STATE \hspace{1em} \textbf{return} $\theta_G$, $\mathcal{L}_{train}$, $\mathcal{A}_{train}$, $\mathcal{P}_{train}$, $\mathcal{R}_{train}$, $\mathcal{F}_{train}$, $\mathcal{L}_{test}$, $\mathcal{A}_{test}$, $\mathcal{P}_{test}$, $\mathcal{R}_{test}$, $\mathcal{F}_{test}$

\end{tcolorbox}

\begin{tcolorbox}[title=Main Algorithm]
\STATE Split into training $(X_{train}, y_{train})$ and testing $(X_{test}, y_{test})$ sets
\STATE Partition $\mathcal{D}_{train}$ into $N$ clients: $\{\mathcal{D}_{train}^i\}$

\STATE Define KAN Classifier
\STATE Initialize global model $\theta_G$

\STATE $\theta_G, \mathcal{L}_{train}, \mathcal{A}_{train}, \mathcal{P}_{train}, \mathcal{R}_{train}, \mathcal{F}_{train}, \\ \mathcal{L}_{test}, \mathcal{A}_{test}, \mathcal{P}_{test}, \mathcal{R}_{test}, \mathcal{F}_{test} \leftarrow \texttt{federated\_learning\_KAN}(\theta_G, \{\mathcal{D}^i\}, \\ \mathcal{D}_{test}, R, S)$
\end{tcolorbox}

\end{algorithmic}
\label{tab:pseudo_code}
%\vspace{-.2cm}
\end{algorithm}

\section{Numerical Results}

\subsection{Simulation Setup}
The experimental setup comprises a federated learning environment with multiple clients, each of which has a subset of a different classification dataset (iris dataset in our example\footnote{Online: https://archive.ics.uci.edu/dataset/53/iris, Available: July 2024.}). The dataset contains features and corresponding labels that come from different sources. The clients train their KAN models locally with the Adam optimizer and a cross-entropy loss function. After local training, the model updates are aggregated by a central server using a federated averaging algorithm. The models are evaluated based on their classification accuracy, precision, recall and F1-score. The code of the F-KAN and F-MLP is available in Github\footnote{Online: https://github.com/ezeydan/F-KANs.git, Available: July 2024.}. Table \ref{tab:sim_params} shows the simulation parameters used in our evaluations.

\begin{table}[h!]
\scriptsize
\caption{Simulation Parameters}
\centering
\begin{tabular}{|c|c|}
\hline
\textbf{Parameter} & \textbf{Value} \\
\hline
\multicolumn{2}{|c|}{\textbf{General Parameters}} \\
\hline
Number of clients ($N$) & 2 \\
Number of rounds ($R$) & 20 \\
Training steps per round ($S$) & 20 \\
Batch size & 16 \\
Learning rate & 0.001 \\
Weight decay & $1 \times 10^{-5}$ \\
Dropout probability & 0.5 \\
\hline
\multicolumn{2}{|c|}{\textbf{F-KAN Parameters}} \\
\hline
Width of layers & [4, 20, 20, 20] \\
Grid size & 5 \\
Spline order ($k$) & 3 \\
Seed & 0 \\
\hline
\multicolumn{2}{|c|}{\textbf{F-MLP Parameters}} \\
\hline
Input size & 4 \\
Hidden layer sizes & [20, 20, 20] \\
Output size & 3 \\
\hline
\end{tabular}
\label{tab:sim_params}
%\vspace{-.5cm}
\end{table}

\subsection{Evaluation Results}

\begin{figure}[htp!]
\begin{subfigure}{\linewidth}
  \centering
\includegraphics[width=.95\linewidth]{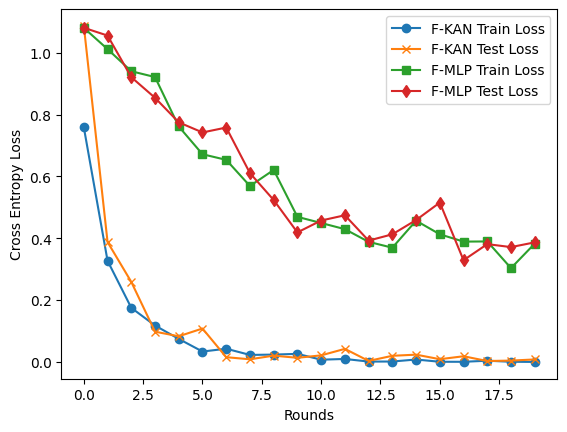}
\caption{}
\label{normal}
\end{subfigure}
\hspace{1mm}
\begin{subfigure}{\linewidth}
  \centering
\includegraphics[width=.95\linewidth]{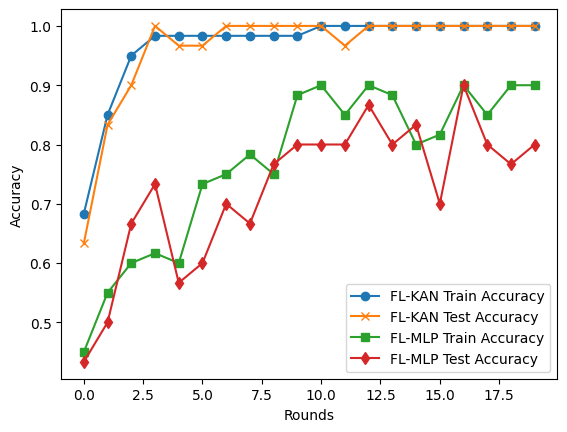}
\caption{}
\label{attack}
\end{subfigure}
\caption{Comparisons of federated learning with F-KAN model and  F-MLP model over rounds (a) Cross-Entropy loss. (b) Accuracy }
\label{fig:FL_MLP}
\vspace{-.5cm}
\end{figure}

Training and evaluation were conducted for both the F-KAN and F-MLP models over 20 rounds of FL. In each round, the model is trained on two client datasets and then the performance is evaluated on a test dataset. From the training information in Fig. \ref{fig:FL_MLP} results in the following comparison between the F-KAN and F-MLP models.  Regarding training and test losses, the F-KAN model shows a rapid decrease in training and test losses, reaching values close to zero in the middle of the training rounds. This indicates efficient learning and excellent generalization abilities. The F-MLP model shows a slower and more variable decline in losses compared to F-KAN. Although it improves over time, losses remain higher and fluctuate more.

\begin{figure*}[htp!]
\begin{subfigure}{.33\linewidth}
  \centering
\includegraphics[width=\linewidth]{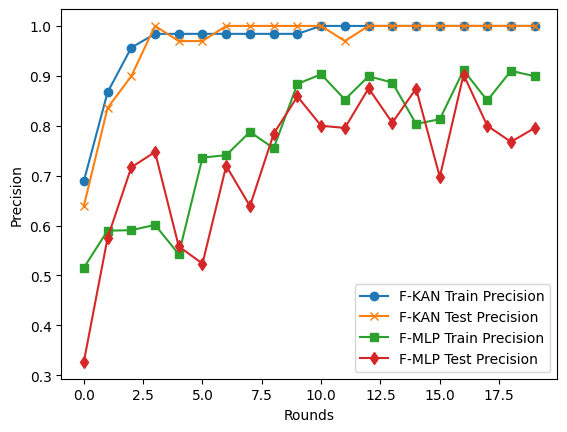}
\caption{}
\label{normal}
\end{subfigure}%
%\hspace{1mm}
\begin{subfigure}{.33\linewidth}
  \centering
\includegraphics[width=\linewidth]{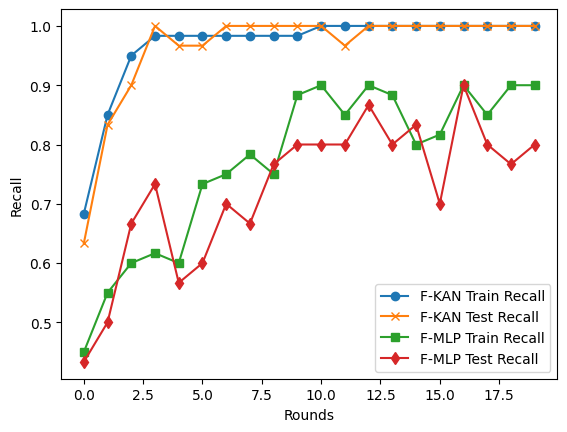}
\caption{}
\label{attack}
\end{subfigure}%
%\hspace{1mm}
\begin{subfigure}{.33\linewidth}
  \centering
\includegraphics[width=\linewidth]{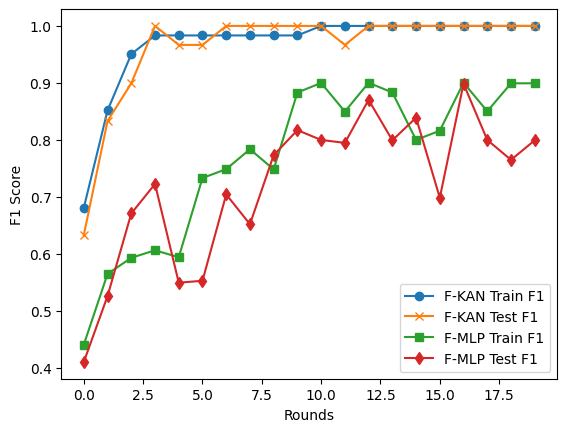}
\caption{}
\label{attack}
\end{subfigure}
\caption{Comparisons of federated learning with F-KAN model and F-MLP model over rounds (a) Precision. (b) Recall. (c) F1-Score }
\label{fig:FL_prec_recall}
\vspace{-.5cm}
\end{figure*}

In terms of accuracy, the F-KAN model achieves 100\% accuracy on both the training and test datasets from round 7 onwards, demonstrating its ability to perfectly classify the data after a few rounds. The F-MLP model achieves moderate accuracy improvements and stabilizes at 90\% for training accuracy and 80\% for test accuracy in the later rounds. However, it does not achieve the same level of perfection as the F-KAN model. In terms of precision, recall and F1 score, these metrics quickly converge to 1.0 as shown in Fig. \ref{fig:FL_prec_recall}, indicating that the F-KAN model is not only accurate but also consistent in its predictions, capturing all true positives without generating false positives or negatives. These metrics for the F-MLP model also improve, but show greater variability. They do not reach the consistently high values as with F-KAN, which indicates a certain instability in performance. The training times for the F-KAN and F-MLP models are observed to be approximately 313.61 seconds and 2.35 seconds respectively.

To summarize, F-KAN shows superior performance with fast convergence, high accuracy and stable precision, recall and F1 scores. The model learns and generalizes effectively across the training and test datasets and achieves perfect scores from round 7 onwards. The F-MLP model, on the other hand, shows improvement but more variability and instability in the performance metrics. While the F-MLP model achieves reasonable accuracy, it does not match the consistency and robustness of the F-KAN model.

%Initial results show that the MLP model can be trained significantly faster than the KAN model. In terms of performance, the KAN model achieves higher test accuracy in some rounds, but is subject to fluctuations, indicating possible overfitting despite regularization. In contrast, the MLP model shows a more stable decrease in test loss and a constant improvement in accuracy, indicating better generalization, albeit with a slightly lower final test accuracy compared to the KAN model.

\subsection{Key Observations}

The rapid convergence and stable performance metrics of the F-KAN model indicate a highly efficient learning process. The spline-based univariate features enable F-KAN to capture complex patterns quickly and accurately, resulting in stable and high performance with fewer training rounds. Although the F-MLP model improves over time, it shows greater variability in performance. This could be due to its reliance on traditional linear weights and activation functions, which may not capture the complexity of the data as well as F-KAN's spline-based approach. The ability of the F-KAN model to achieve and maintain high accuracy precision, recall, and F1 score on both the training and test datasets in relatively early rounds indicates excellent generalization. This is crucial for applications where consistent and reliable performance on unseen data is critical, such as medical diagnosis or autonomous driving.  The F-MLP model shows adequate but less consistent generalization capabilities. Its fluctuating performance metrics suggest that it is more prone to overfitting or underfitting, making it less reliable for applications that require high accuracy and consistency.

The F-KAN model takes approximately 301.19 seconds to train, reflecting the complexity of the model and the computational cost of the spline-based functions. However, these costs are justified by the superior performance and stability achieved. The F-MLP model trains much faster (1.83 seconds), which could be advantageous in scenarios where fast model updates are required, such as in real-time systems. However, the trade-off is lower and more variable performance. The high accuracy, precision, recall, and F1 scores make the F-KAN ideal for applications where errors can have significant consequences, e.g. in critical infrastructure monitoring. F-KAN's ability to capture complex patterns with fewer parameters makes it suitable for tasks involving complicated data relationships, such as image and speech recognition. In addition, the shorter training time makes F-MLP suitable for applications where models need to be updated quickly and frequently, such as recommendation systems or real-time analysis. F-MLPs may be preferred in scenarios where computing resources are limited and a slight compromise in performance in favour of faster processing times is acceptable.

\section{F-KANs for Wireless Case Studies}

The proposed F-KANs can be an example of a distributed AI approach that meets the need for decentralized, adaptive intelligence in native networks. This approach can provide several advantages for distributed AI in wireless networks:
\begin{itemize}[leftmargin=*]
    \item In contrast to traditional neural networks, which are based on fixed activation functions, F-KANs use spline-based univariate functions as learnable activation functions, which allow for better adaptability and interpretability. Thanks to this adaptability, F-KANs can learn complex patterns in data more effectively, which is particularly important for the diverse and dynamic conditions in wireless networks.
    \item The federated architecture of F-KANs enables decentralized model training across multiple clients without the need to transfer raw data, thereby preserving user privacy. This is achieved by performing local computations on each device and aggregating model updates on a central server, which then updates the global model. Such a system meets the privacy and security requirements of modern wireless networks.
    \item  F-KANs have shown superior performance on classification tasks compared to traditional federated models such as F-MLPs. The use of univariate features enables fast convergence, high accuracy and stable performance metrics such as precision, recall and F1-score. These features are crucial for applications in wireless/native networks that require reliable and timely predictions since they have already AI deeply embedded into their core functionalities and operations.
\end{itemize}

\subsection{Challenges and Open Research Areas}

While distributed AI frameworks like F-KANs offer promising solutions for native networks, several challenges and research opportunities exist:

\begin{itemize}[leftmargin=*]
    \item Despite the advantages of federated learning, the exchange of model updates between clients and the central server can introduce significant communication overhead especially for bigger models. For F-KANs, optimizing the frequency and size of model updates is going to be crucial to maintain a balance between performance and bandwidth consumption. Techniques such as model compression, update sparsification, and adaptive communication protocols are active areas of research that can be used to improve the efficiency of F-KAN deployment.
    \item F-KAN agents in native networks need to communicate effectively across different network layers and paradigms. Optimizing these communication protocols to support efficient and low-latency information exchange is crucial, especially in scenarios requiring rapid decision-making and response, such as autonomous vehicles, drone networks, and smart grid applications.
    \item The spline-based activation functions in F-KANs are highly flexible, allowing for improved learning. However, the computational complexity of F-KAN models can pose challenges in environments with limited computational and energy resources, such as edge devices and sensor networks. Future research needs to balance model complexity and resource constraints by designing lightweight, efficient AI models that can be deployed across diverse network nodes.
    \item Native networks are dynamic by nature, with fluctuating traffic loads, device availability and possible failures. The decentralized nature of F-KANs makes them well suited for environments with dynamic connectivity, such as mobile networks or sensor arrays. However, fluctuating network conditions, device availability and data heterogeneity can impact the convergence and performance of the model. F-KAN models must withstand these dynamics and ensure reliable performance even when parts of the network are disrupted. This requires the development of F-KANs with adaptive training strategies and fault-tolerant algorithms capable of handling non-IID (Independent and Identically Distributed) data distributions and network partitions to ensure consistent performance in highly variable network environments.
    \item While F-KANs offer inherent privacy benefits due to their federated learning structure, the decentralized nature of these networks can introduce vulnerabilities such as model inversion attacks or poisoning attacks. Exploring secure aggregation techniques, differential privacy mechanisms and robust model verification for F-KANs can further strengthen their resilience against adversarial threats.
\end{itemize}

\subsection{Future Directions  for F-KANs in Native Networks}

To fully realize the potential of distributed AI in native networks, future research needs to address the above challenges while exploring new use cases and applications. Some important directions are:

\begin{itemize}[leftmargin=*]
    \item A promising area of future research is the development of dynamic update strategies for F-KANs that adjust communication frequency and model aggregation based on network conditions. This would improve the efficiency of training in environments with intermittent connectivity or varying bandwidth availability and ensure optimal utilisation of resources while maintaining high performance.
    \item The use of multi-agent reinforcement learning within the F-KAN framework can improve the decision-making processes of distributed agents in the network. By enabling F-KAN models to learn optimal strategies for managing network resources, balancing traffic load, and adapting to environmental changes, MARL has the potential to improve both the efficiency and intelligence of the network as a whole.
    \item The modular nature of F-KANs allows for potential hybridization with other AI architectures tailored to specific applications within native networks. For example, combining F-KANs with convolutional neural networks (CNNs) for image-based network tasks or with graph neural networks (GNNs) for topology-aware resource management can lead to specialized models that are both efficient and effective.
    \item F-KAN research can be accelerated by establishing open source platforms for sharing model architectures, source code and standardized datasets. Such collaboration would support reproducibility, enable benchmarking across different wireless network scenarios, and foster the development of robust and adaptable distributed AI frameworks.
\end{itemize}

Finally, Table \ref{tab:fkan-vs-fmlp} provides a comparison of F-KAN and F-MLP for use cases in network resource management, edge computing/IoT, network security and privacy, and autonomous network control, highlighting their characteristics, advantages and disadvantages in B5G/6G cellular networks.

\begin{table*}[ht]
\footnotesize
\centering
\caption{Comparison of F-KAN and F-MLP Across Use Cases and Domains in B5G/6G Wireless Networks}
\begin{tabular}{|p{2.5cm}|p{2.5cm}|p{3.0cm}|p{2.5cm}|p{5.5cm}|}
\hline
\textbf{Use Case / Domain}          & \textbf{Characteristic}  & \textbf{F-KAN Approach} & \textbf{F-MLP Approach} & \textbf{Pros and Cons} \\ \hline
\multirow{3}{*}{\begin{tabular}[l]{@{}l@{}} \textbf{Network Resource} \\ \textbf{Management}   \\ \end{tabular} } & Scalability & \begin{tabular}[l]{@{}l@{}} High due to spline-\\based functions   \\ \end{tabular}  & Moderate, limited by linear activations & 
\begin{tabular}[c]{@{}l@{}}F-KAN is better suited for large-scale \\ networks with dynamic resource demands. \\ F-MLP may face scalability bottlenecks.\end{tabular} \\ \cline{2-5} 
                                          & Adaptability    & \begin{tabular}[c]{@{}c@{}}\begin{tabular}[l]{@{}l@{}} Real-time, quick \\ response to \\ network changes   \\ \end{tabular} 
                                          \end{tabular} & \begin{tabular}[c]{@{}c@{}}Slower adaptation,\\ fixed activations\end{tabular} & 
\begin{tabular}[c]{@{}l@{}}F-KAN achieves faster and more precise\\ adaptability, essential for dynamic \\ network slices in 6G. F-MLP slower to adapt.\end{tabular} \\ \cline{2-5} 
                                          & Privacy Preservation & \begin{tabular}[c]{@{}c@{}}\begin{tabular}[l]{@{}l@{}} High through \\ federated learning, \\ localized model training   \\ \end{tabular} 
 \end{tabular} & Similar, but less optimized for FL & 
\begin{tabular}[c]{@{}l@{}}Both approaches ensure privacy, \\ but F-KAN offers better performance\\ under privacy constraints.\end{tabular} \\ \hline
\multirow{3}{*}{ \begin{tabular}[l]{@{}l@{}} \textbf{Edge Computing/} \\ \textbf{IoT Networks}   \\ \end{tabular}  } & Efficiency & \begin{tabular}[c]{@{}c@{}} \begin{tabular}[l]{@{}l@{}} High through \\ FL localized model \\ training   \\ \end{tabular} \end{tabular} & Low computational complexity & 
\begin{tabular}[c]{@{}l@{}}F-KAN’s computational overhead is higher,\\ but improved performance justifies the cost\\ in critical IoT applications. F-MLP is faster, \\ suitable for low-resource nodes.\end{tabular} \\ \cline{2-5} 
                                          & Energy Consumption & \begin{tabular}[l]{@{}l@{}}Energy-efficient \\ model training\\ but requires more \\ computational power\end{tabular} & \begin{tabular}[c]{@{}c@{}}Low energy, less\\ computational demand\end{tabular} & 
\begin{tabular}[c]{@{}l@{}}F-KAN achieves better accuracy at a\\ slight energy cost. F-MLP is more\\ suitable for energy-constrained environments.\end{tabular} \\ \cline{2-5} 
                                          & Generalization    & \begin{tabular}[l]{@{}l@{}}High accuracy and \\ stable metrics\end{tabular}  & Moderate accuracy and variance & 
\begin{tabular}[c]{@{}l@{}}F-KAN excels in handling diverse data from \\ edge devices. F-MLP may struggle with \\ heterogeneous data distributions.\end{tabular} \\ \hline
\multirow{3}{*}{\begin{tabular}[l]{@{}l@{}} \textbf{Network Security} \\ \textbf{and Privacy}   \\ \end{tabular} } & Attack Resilience & \begin{tabular}[l]{@{}l@{}}Better resistance \\ to adversarial\\ attacks via complex \\ functions\end{tabular} & Moderate resilience & 
\begin{tabular}[c]{@{}l@{}}F-KAN’s spline functions increase \\ model robustness, enhancing security.\\ F-MLP is more prone to model attacks.\end{tabular} \\ \cline{2-5} 
                                          & Privacy Optimization & \begin{tabular}[c]{@{}c@{}}Strong privacy by local \\ model training with FL\end{tabular} & Standard FL, less optimal & 
\begin{tabular}[c]{@{}l@{}}F-KAN provides higher performance \\ under privacy constraints; F-MLP is less \\ capable of optimizing privacy-performance \\ trade-offs.\end{tabular} \\ \cline{2-5} 
                                          & Complexity & \begin{tabular}[l]{@{}l@{}}Increased due to \\ complex spline-based\\ activation functions\end{tabular} & Simpler linear architecture & 
\begin{tabular}[c]{@{}l@{}}F-KAN’s complexity can be a drawback \\ for real-time security applications,\\ while F-MLP’s simplicity makes it \\ easier to deploy quickly.\end{tabular} \\ \hline
\multirow{3}{*}{\begin{tabular}[l]{@{}l@{}} \textbf{Autonomous Network } \\ \textbf{Control}   \\ \end{tabular} } & Learning Speed & Fast convergence, early accuracy gains & \begin{tabular}[c]{@{}c@{}}Slower convergence,\\ longer training times\end{tabular} & 
\begin{tabular}[c]{@{}l@{}}F-KAN quickly reaches high accuracy,\\ beneficial for autonomous control;\\ F-MLP requires longer training.\end{tabular} \\ \cline{2-5} 
                                          & Real-Time Adaptation & High, supports real-time decision-making & Limited due to slower learning & 
\begin{tabular}[c]{@{}l@{}}F-KAN’s rapid learning supports \\ autonomous decision-making in real-time\\ applications (e.g., UAV control),\\ while F-MLP is less responsive.\end{tabular} \\ \cline{2-5} 
                                          & Model Complexity & Moderate-high, spline function usage & Low, linear weight-based & 
\begin{tabular}[c]{@{}l@{}}F-KAN’s complexity improves \\ decision-making in complex tasks.\\ F-MLP is simpler, more applicable for \\ tasks requiring quick deployment.\end{tabular} \\ \hline
\end{tabular}
\label{tab:fkan-vs-fmlp}
\end{table*}

\section{Conclusions and Future Work}

This paper shows the effectiveness of federated KANs in classification tasks. F-KANs combine the strengths of federated learning with the innovative architecture of F-KANs to achieve higher accuracy with fewer parameters while preserving data privacy. The results highlight the potential of F-KANs as a powerful tool for classification in decentralized environments. Overall, the F-KAN model proved to be more reliable and effective for the FL task under consideration in comparison to federated F-MLP model, as shown by its consistently high performance across metrics and rounds. On the other hand, the computational costs of spline-based functions must be better optimized. We have also highlighted F-KANs as a wireless case study. Future work will focus on further optimizing the F-KAN architecture and investigating its applicability to other classification tasks in federated environments, e.g.  image classification in satellite and edge networks.

%\section*{Acknowledgment}

\balance

%\vspace{12pt}

\end{document}